\documentclass[10pt,twocolumn,letterpaper]{article}

\usepackage{cvpr}              

\usepackage{graphicx}
\usepackage{amsmath}
\usepackage{amssymb}
\usepackage{booktabs}
\usepackage{bbm}
\usepackage{multirow}
\usepackage{color}
\usepackage[labelformat=simple]{subcaption}

\graphicspath{ {./images/} }

\usepackage[pagebackref,breaklinks,colorlinks]{hyperref}

\usepackage[capitalize]{cleveref}
\crefname{section}{Sec.}{Secs.}
\Crefname{section}{Section}{Sections}
\Crefname{table}{Table}{Tables}
\crefname{table}{Tab.}{Tabs.}

\def\cvprPaperID{5596} 
\def\confName{CVPR}
\def\confYear{2023}

\begin{document}

\title{CascadeV-Det: Cascade Point Voting for 3D Object Detection}

\author{Yingping Liang\textsuperscript{$1$} \qquad \qquad  Ying Fu\textsuperscript{$1$}\thanks{Corresponding Author: fuying@bit.edu.cn} \\
	\textsuperscript{$1$}Beijing Institute of Technology \\ 
}

\maketitle
\begin{abstract}
    Anchor-free object detectors are highly efficient in performing point-based prediction without the need for extra post-processing of anchors. However, different from the 2D grids, the 3D points used in these detectors are often far from the ground truth center, making it challenging to accurately regress the bounding boxes. To address this issue, we propose a Cascade Voting (CascadeV) strategy that provides high-quality 3D object detection with point-based prediction. Specifically, CascadeV performs cascade detection using a novel Cascade Voting decoder that combines two new components: Instance Aware Voting (IA-Voting) and a Cascade Point Assignment (CPA) module. The IA-Voting module updates the object features of updated proposal points within the bounding box using conditional inverse distance weighting. This approach prevents features from being aggregated outside the instance and helps improve the accuracy of object detection. Additionally, since model training can suffer from a lack of proposal points with high centerness, we have developed the CPA module to narrow down the positive assignment threshold with cascade stages. This approach relaxes the dependence on proposal centerness in the early stages while ensuring an ample quantity of positives with high centerness in the later stages. Experiments show that FCAF3D with our CascadeV achieves state-of-the-art 3D object detection results with 70.4\% mAP@0.25 and 51.6\% mAP@0.5 on SUN RGB-D and competitive results on ScanNet.
\end{abstract}

In this paper, we first point out the essential difference between anchor-free 2D and 3D object detection, which is actually how to define positive and negative
training samples, which leads to the performance gap between them.

\begin{figure}[t]
    \centering
    \subcaptionbox{\label{2} Voting and Prediction \cite{qi2019deep}}{\includegraphics[width = .5\linewidth]{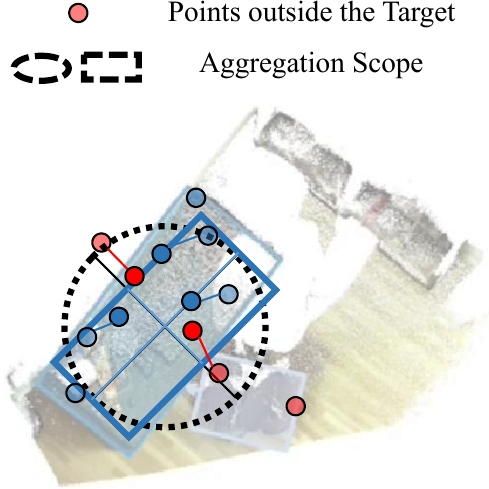}}\hfill
    \subcaptionbox{\label{1} Per-Point Prediction \cite{rukhovich2021fcaf3d}}{\includegraphics[width = .5\linewidth]{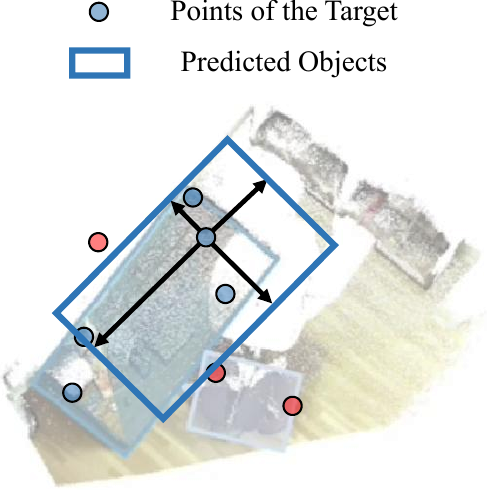}}
    \subcaptionbox{\label{3} Our Cascade Voting and Prediction}{\includegraphics[width = 1.0\linewidth]{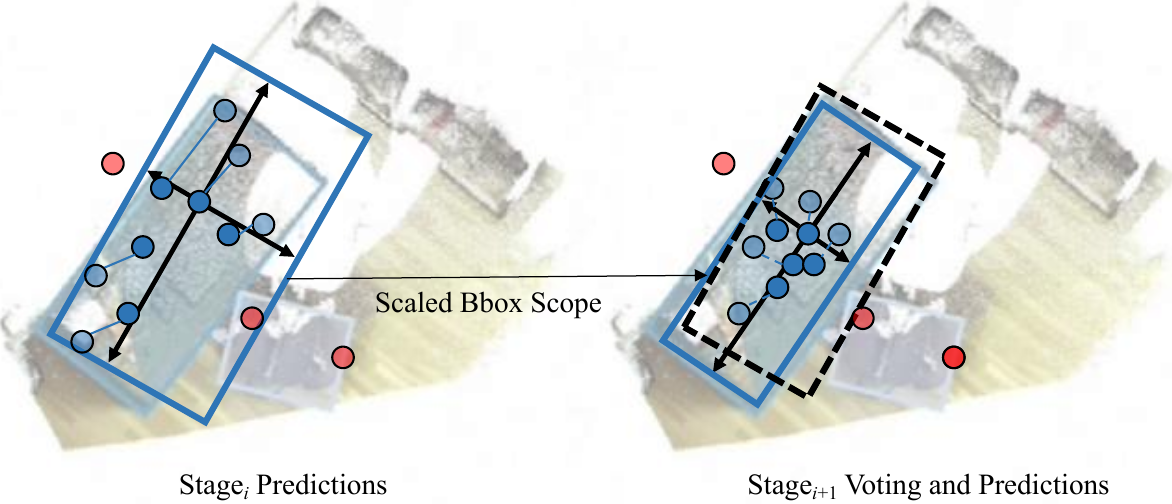}}
    \caption{Points in red are the harmful object features outside the real target and the point colors from light to dark indicate the direction to vote. All points in the black ball of voting are assigned to derive the object features, resulting in aggregated features outside the real instance. Per-point prediction can also be hard to regress accurately when directly predicting bounding boxes with points far from the ground truth centers. Our method allows us to update the proposal points and object features in a cascade voting manner trained with abundant high quality positives. For clarity, we show the methods from a BEV perspective.}
    \label{fig:denoising}
\end{figure}

\section{Introduction}
\label{sec:intro}


3D object detection from point clouds is an essential task in 3D scene understanding, with promising applications in robotics \cite{wang2019densefusion, duan2021robotics, pires2022obstacle} and augmented reality \cite{kung2016efficient, liu2020learning}. And many recently proposed 3D object detectors with point clouds are built based on the anchor-free framework \cite{rukhovich2021fcaf3d, wang2021fcos3d, vu2022softgroup, wang2021centernet3d} due to its efficiency and simple design, where bounding boxes regression is framed as the point-based prediction. Among them, voting \cite{qi2019deep, qi2020imvotenet, yang2022boosting, vu2022softgroup} performs the crucial feature grouping step, which votes to object centers and then aggregates the features. But voting methods may also fail in cluttered scenes with edge points where various points are close but belong to different objects or empty spaces as shown in Figure \ref{2}, leading to feature-wise noise. Per-point prediction methods \cite{wang2021centernet3d, rukhovich2021fcaf3d, tian2022fully} are then proposed to mitigate this and avoid any prior assumptions about objects. However, 3D object centers are likely to be in empty space and far away from any point, making it hard to aggregate object features in the object centers as shown in Figure \ref{1}. 



To more accurately estimate the ground truth center and predict bounding boxes, we present a Cascade Voting Detector (CascadeV-Det) with an extra multi-stage decoder. The decoder consists of cascade stages combining a novel Instance Aware Voting (IA-Voting) module for feature updating and trained with a Cascade Point Assignment (CPA) strategy for positive assignment, which bridges the gap between voting methods and per-point prediction methods.

To be specific, the decoder updates the proposal points with the centers of the predicted bounding boxes to reduce the distance to the object centers. To further update the object features corresponding to the updated proposal points, the IA-Voting module aggregates the object features within the instances using conditional inverse distance weighted and thus avoids aggregating object features from unexpected areas. And the instances can be represented by class-agnostic bounding boxes that are predicted by the detection head from the former stage.

In addition, using only points with high centerness as positives for training can improve the quality of prediction but results in few proposal points matched as positives, making it hard to train the model. Thus we propose the CPA strategy, which trains the decoder stages sequentially with progressively decreasing thresholds. Therefore, the stages deeper into the decoder still get abundant positives and are more selective against close false positives. In addition, to avoid any mismatched ground truths, the CPA also additionally feeds ground truth centers with noises as fixed positives and trains the model to reconstruct the original boxes.


Besides, images can be used intuitively as auxiliary information for 3D object detection to provide geometric and semantic cues. By introducing deformable attention with deformable references in the transformer layers, we modify the cascade decoder to support and aggregate both image features and point features. And the more accurate positioning of the ground truth center from CascadeV-Det also provides more precise image reference localization.

We validate the effectiveness of our method with extensive experiments with multiple metrics on SUN RGB-D \cite{song2015sun} and ScanNet \cite{dai2017scannet}. Our CascadeV-Det surpasses the state-of-the-art methods by at least 2.8\% mAP@0.25 and yields a significant improvement.

\begin{figure*}[t]
   \centering
   \includegraphics[width=1.0\textwidth]{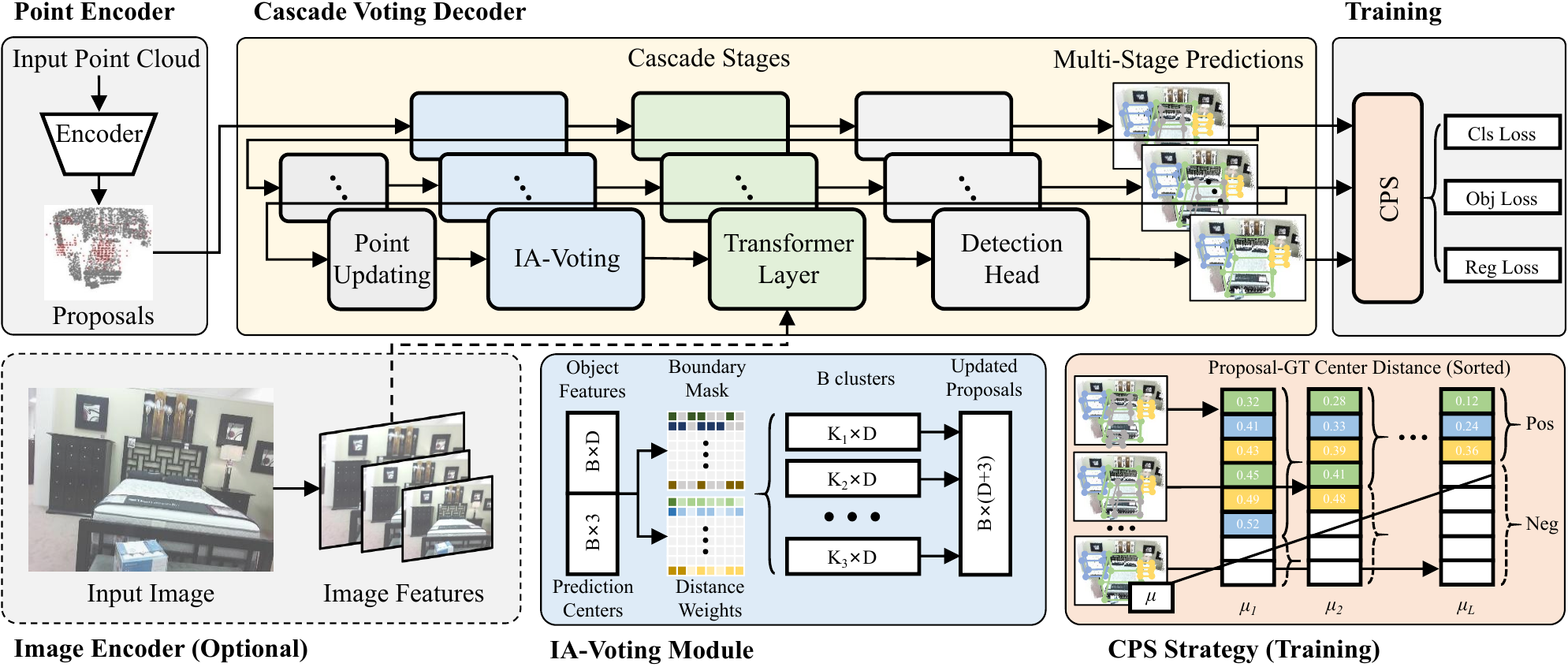}
   \caption{The framework of CascadeV-Det with a point encoder and a novel cascade voting decoder with IA-Voting modules. An extra CPA strategy is also used for training. Proposal points are first selected from the point encoder. Then the object features are updated by the IA-Voting module and fed into transformer layers with attention modules for feature refinement with per-stage predictions from the detection heads. The threshold for positives in the CPA strategy decreases stage by stage, providing stricter positive supervision with abundant high quality positives in the deeper stages.}
   \label{fig:framework-simple}
\end{figure*}

We summarize our contributions as follows.

\begin{itemize}
\item We propose a novel 3D object detector named CascadeV-Det to provide high quality detections by combining a feed-forward proposal updating process and a cascade training strategy.
\item We design an IA-Voting module for updating object features with updated proposal points to precisely aggregate instance-level object features.
\item We design a CPA strategy that formulates a cascade positive assignment process and trains a sequence of cascade stages to provide abundant positives with high centerness by decreasing the assignment threshold.
\end{itemize}

\section{Related Work}

\noindent \textbf{3D object detection with point clouds.} Point cloud representation is a common data format for 3D scene understanding. To locate and recognize 3D objects using purely geometric information, early methods \cite{yang2018pixor, zhou2018voxelnet, deng2021voxel} use 2D projection or voxels to avoid expensive 3D operations. For example, Pixor \cite{yang2018pixor} obtains the 2D BEV map with the height and reflectance as channels, and then the fine-tuned RetinaNet \cite{lin2017focal} is used for object detection. And VoxelNet \cite{zhou2018voxelnet} uses stacked encoding layers to extract voxel features. However, these methods suffer from large memory and computational cost when inputting large scenes. Grouping methods \cite{qi2020imvotenet, qi2019deep, vu2022softgroup} are then proposed to solve this problem. For example, VoteNet \cite{qi2019deep} uses Hough Voting to detect boxes by features sampling, grouping, and voting operations designed for 3D data with ball query and a generic point set learning network to aggregate the votes, which fails in clustered scenes. And RoI-Conv \cite{wang2022cagroup3d} pooling is also proposed to aggregate specific features, but focuses only on local regions of fixed size and ignores global correlations. To avoid grouping operation, FCAF3D \cite{rukhovich2021fcaf3d} performs fully convolutional anchor-free 3D object detection with proposal point bounding box parametrization but may fail to locate the geometrically irregular objects due to coarse proposal points with distance noise and thus mismatched ground truth center. Our CascadeV-Det detects the objects with a cascade decoder where proposal points and object features are both well-updated and refined step-by-step closer to the ground truth centers, leading to high quality detections. 

~\\
\noindent \textbf{3D object detection with RGB-D data.} Notably, current 3D detection methods have achieved great success with only geometric input. To further boost detection performance, leveraging RGB information is a potential direction. Some methods guide the search space in 3D with object proposals from the image features. For example, EPNet \cite{huang2020epnet} and EPNet++ \cite{liu2021epnet++} fuse image features with point features from the backbone’s intermediate layers. Other 3D-driven methods first extract 3D objects and then output predictions using the 2D features. For example, ImVoteNet \cite{qi2020imvotenet} encodes 2D detection results to guide the voting operations in VoteNet, focusing more on fusing 2D and 3D features. And DeMF \cite{yang2022boosting} adaptively aggregates image features by taking the projected 2D location of the 3D point as a reference, but neglects the refinement of 2D features by 3D geometric information. Our method introduces deformable attention into the decoder where queries are refined alternately with both point and image features by the attention modules.

\section{The Proposed CascadeV-Det}
We propose a cascade 3D object detector named CascadeV-Det. As shown in Figure \ref{fig:framework-simple}, the CascadeV-Det consists of a point encoder and a Cascade Voting decoder with IA-Voting modules and is trained with CPA strategy. In this section, after a summary of the prediction formulation and method motivation in Section \ref{fcaf3d}, we introduce our proposed IA-Voting module to perform the object feature updating process in Section \ref{mqr-section}. Then, we describe how the CPA strategy performs positive assignment in a cascade manner with decreasing threshold and increasing proposal point centerness in Section \ref{CPA}. Finally, we explain how the image features are fused to support the 3D detection by deformable attention in the transformer layers in Section \ref{img}. 

\subsection{Formulation and Motivation}
\label{fcaf3d}

Our CascadeV-Det uses a point encoder with the point cloud input to provide proposals and a decoder to predict objects. To be specific, we feed the sampled points and object features from the point encoder as the proposals into the decoder, which contains cascade stages each consisting of an IA-Voting module, a transformer layer, and a detection head. Due to the superior performance of the transformer for multimodal data, we also support the image features from the image encoder as extra input for the transformer layers. Each stage performs proposal point and feature updating and outputs the refined object predictions that are also treated as proposals for the next stage. During the training process, the predicted objects from the detection head of each stage are assigned positives and negatives by the CPA strategy with decreasing thresholds.

\noindent \textbf{Point Encoder.} To be specific, a point encoder works as a simple yet effective backbone following \cite{rukhovich2021fcaf3d}. Given a set of $N$ points $\boldsymbol{P}\in\mathbb{R}^{N\times3}$ associated with its 3-dimensional XYZ coordinates, the point encoder encodes the $N$ points and outputs the point features $\boldsymbol{x}^{p}\in\mathbb{R}^{N\times D}$.

~\\
\noindent \textbf{Cascade Voting Decoder.} Dense and coarse points bring noisy object predictions and extensive false positives. Thus we select the object features $\boldsymbol{q}\in\mathbb{R}^{B\times D}$ from overall point features $\boldsymbol{x}^{p}$ with top $B$ maximum centerness and the corresponding points $\boldsymbol{p}\in\mathbb{R}^{B\times 3}$ as proposals for fine object predictions with an extra Cascade Voting decoder. Each object feature is responsible for predicting the regression value from the corresponding proposal point to the bounding box.

With proposal points and object features, the detection head predicts the objects in the form of classification probabilities $\hat{p}$, bounding box regression parameters $\boldsymbol{\delta}=\{\delta_{1},\dots,\delta_{6}\}$, heading angle $\hat{\theta}$, and centerness $\hat{c}$. The detection is framed as a point-based learning problem as \cite{rukhovich2021fcaf3d, tian2019fcos, duan2019centernet, wang2021centernet3d}, where center distance matching is used to select positive and negative points for training. For a point $\boldsymbol{p}_{i}=(\hat{x},\hat{y},\hat{z})$ and the matched ground truth box $\textbf{g}=(x, y, z, w, l, h, \theta)$, the bounding box regression target $\boldsymbol{\delta}^{*}$ can be defined as:
\begin{equation}
    \begin{aligned}
    \delta_{1}^{*} &=x+\frac{w}{2}-\hat{x}, \delta_{2}^{*}=\hat{x}-x+\frac{w}{2}, \delta_{3}^{*}=y+\frac{l}{2}-\hat{y}, \\
    \delta_{4}^{*} &=\hat{y}-y+\frac{l}{2}, \delta_{5}^{*}=z+\frac{h}{2}-\hat{z}, \delta_{6}^{*}=\hat{z}-z+\frac{h}{2},
    \end{aligned}
\end{equation}
and the centerness depicts the normalized distance from the location to the center of the object that the location is responsible for \cite{tian2019fcos}. A point with higher centerness indicates that it is closer to the ground truth center. Given the regression target $\boldsymbol{\delta}^{*}$, the centerness target is then defined as:
\begin{equation}
    \hat{c}^{*}=\sqrt{\frac{\min \left(\delta_{1}^{*}, \delta_{2}^{*}\right)}{\max \left(\delta_{1}^{*}, \delta_{2}^{*}\right)} \times \frac{\min \left(\delta_{3}^{*}, \delta_{4}^{*}\right)}{\max \left(\delta_{3}^{*}, \delta_{4}^{*}\right) } \times \frac{\min \left(\delta_{5}^{*}, \delta_{6}^{*}\right)}{\max \left(\delta_{5}^{*}, \delta_{6}^{*}\right) }},
\end{equation} 

\begin{figure}[t]
   \centering
   \includegraphics[width=0.45\textwidth]{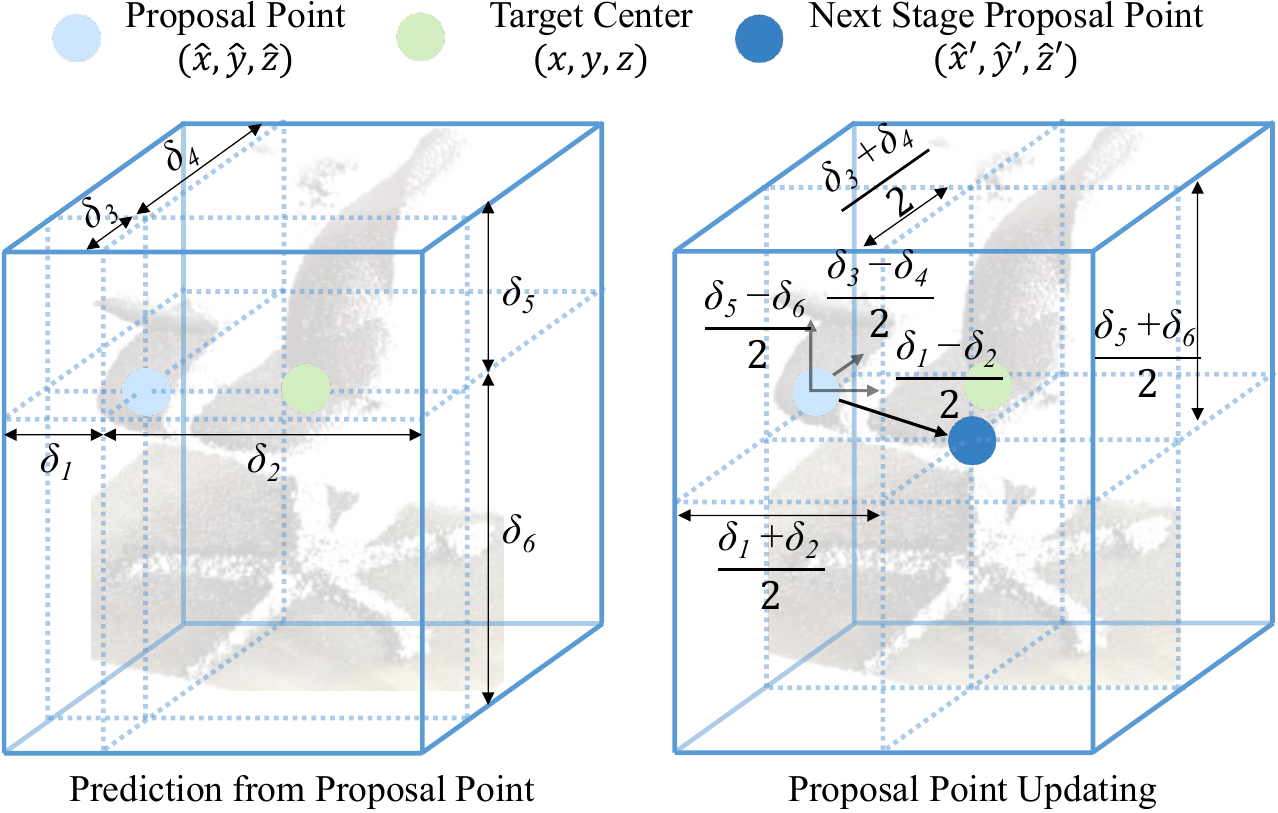}
    \caption{The left shows the bounding box from a single proposal point with predicted $\boldsymbol{\delta}$. The right shows the updating process from the proposal point (gray) to the bounding box center (blue). And the center of the predicted bounding boxes are generally closer to the ground truth centers than the base proposal points.}
    \label{fig:updating}
\end{figure}

In order to provide proposal points closer to the ground truth centers, as shown in Figure \ref{fig:updating}, the proposal points in each stage are updated as the centers of predicted bounding boxes from the detection head of the former stage:
\begin{equation}
    \hat{x}^{\prime}=\hat{x}+\frac{\delta_{1}-\delta_{2}}{2}, \hat{y}^{\prime}=\hat{y}+\frac{\delta_{3}-\delta_{4}}{2}, \hat{z}^{\prime}=\hat{z}+\frac{\delta_{5}-\delta_{6}}{2}.
\end{equation}
However, simply changing the position of the points without updating the object features leads to false predictions. Therefore, we propose the IA-Voting module, which votes to the updated point positions by conditional inverse distance weighting to aggregate and update the object features. Then in the transformer layer, cross-attention and mixed-attention are used to refine the object features with the help of point features and image features, respectively. The detection head is then used to predict objects with refined features and updated proposal points. 

~\\
\noindent \textbf{Model Training.} To train the model, we need to assign the predicted objects to ground truth objects to obtain regression, centerness, and classification targets. Typically if a proposal point $\boldsymbol{p}_{i}=(\hat{x},\hat{y},\hat{z})$ falls into any real ground truth boxes, it is considered a positive matching a ground truth object \cite{rukhovich2021fcaf3d, wang2021fcos3d, qi2020imvotenet}. This is equivalent to defining a scaling threshold $\mu=0.5$. For an axis-aligned ground truth $(x, y, z, w, l, h)$, the corresponding positive satisfies: 
\begin{equation}
    \begin{array}{l}
    x-\mu w \leq \hat{x} \leq x+\mu w, \\
    y-\mu l \ \ \leq \hat{y} \leq y+\mu l, \\
    z-\mu h \ \leq \hat{z} \leq z+\mu h, 
    \end{array}
    \label{eq3}
\end{equation}
where a smaller $\mu$ works well to suppress the false positives since the points far from the ground truth center are treated as negatives and only close enough points with tiny distance noise contribute to the final predictions. However, it is not feasible to decrease the threshold directly, because few proposal points meet the demand, resulting in few positives and an empty prediction.

To address this problem, we propose the CPA strategy, which decreases the positive assignment threshold $\mu$ stage by stage. This is motivated by our observation that the predicted object centers are generally closer to the ground truth centers than the proposal points, which indicates that decreasing threshold $\mu$ at each stage can avoid a sharp reduction in the number of positives. Meanwhile, at the early training, the CPA adds updating points with minimum distance noise from the ground truth centers to prevent any unmatched ground truths.

\begin{figure*}[t]
   \centering
    \subcaptionbox{\label{cas-baseline} Standard Positive Matching \cite{rukhovich2021fcaf3d}}{\includegraphics[width=0.24\textwidth]{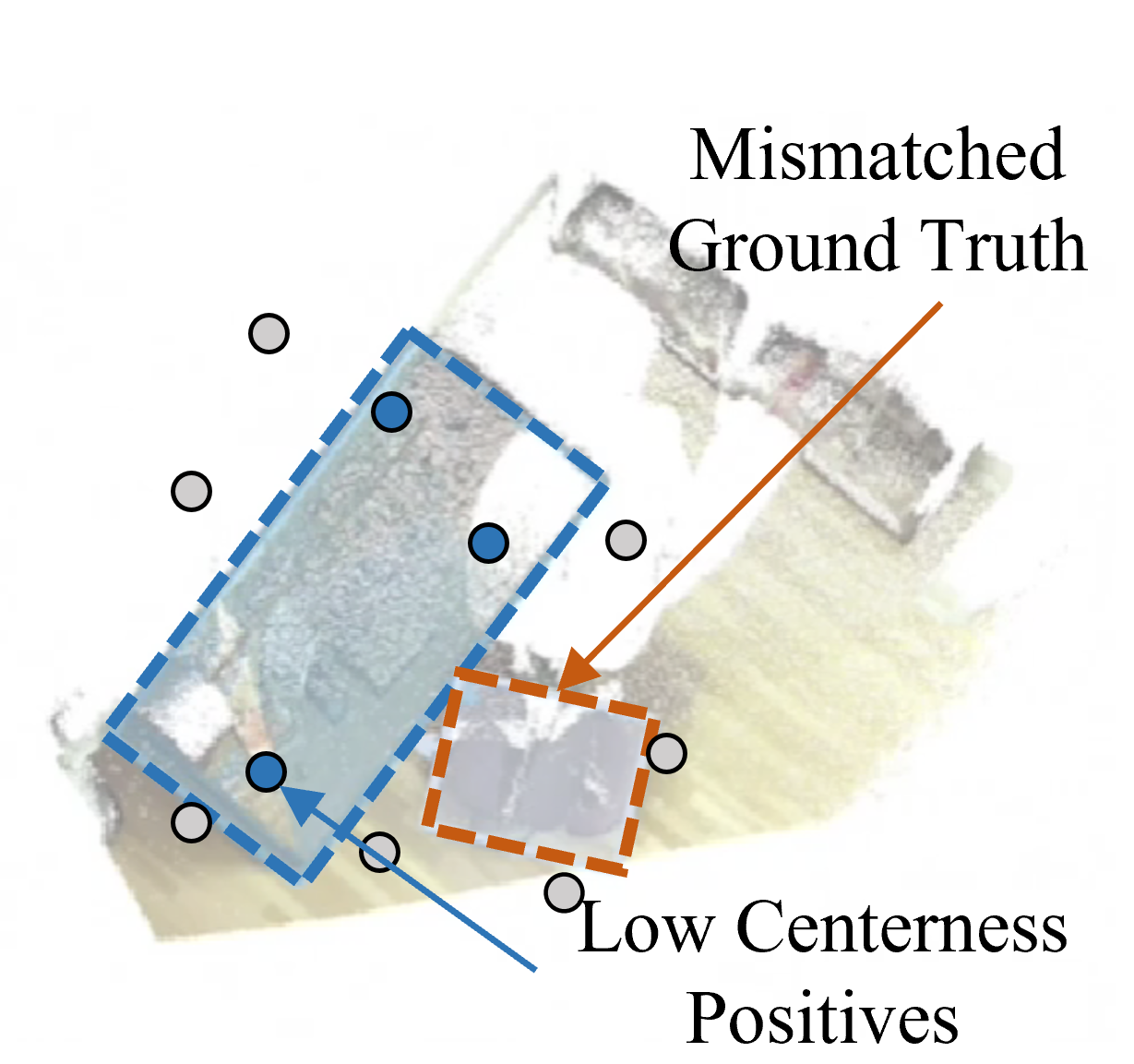}}\hfill
    \subcaptionbox{\label{cas-ours} CPA Strategy with Decreasing Threshold.}{\includegraphics[width=0.72\textwidth]{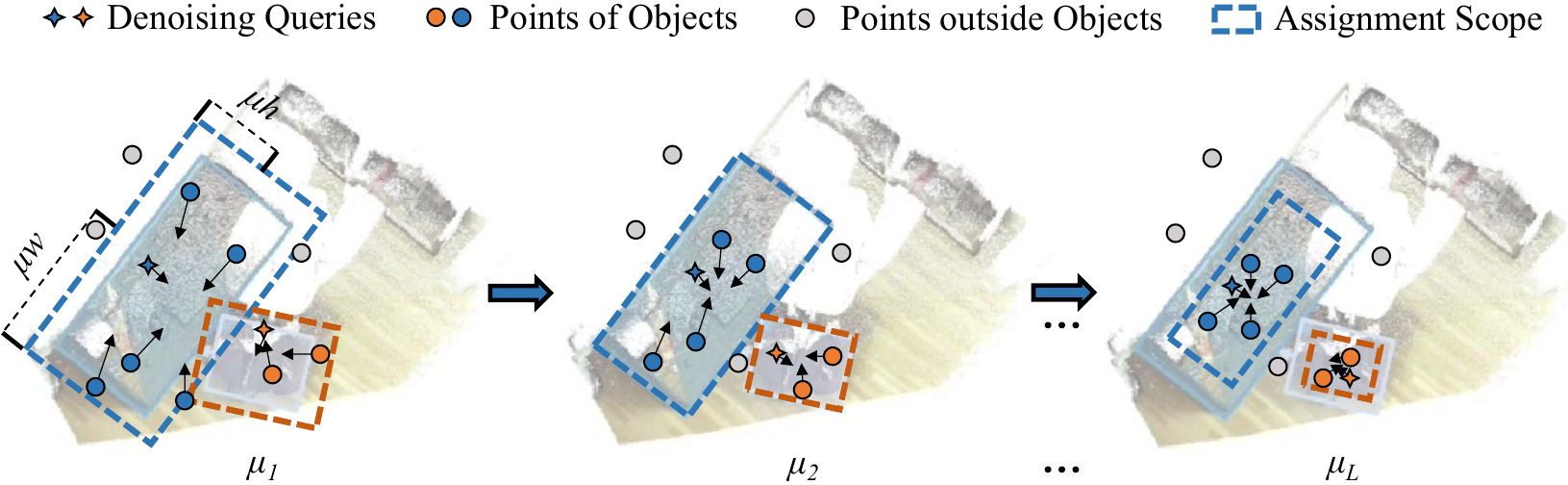}}
   \caption{The matched positive samples in the training process, in which the unmatched proposal points are represented by gray. The stars represent the denoising points (queries) with the minimum distance from the ground truth centers for denoising training. And the arrow direction indicates the updated location of the proposed point. The dashed box indicates the threshold range selected by the positive sample. Note that, unlike IA-Voting, this matching strategy only takes effect during training and is used to assign targets to proposal points.}
   \label{fig:stage-positive}
\end{figure*}

\subsection{Instance-Aware Voting}
\label{mqr-section}

Since we update the position of the proposal points in the decoder, the object features need to be updated accordingly. To do this, we propose the IA-Voting module, which directly aggregates object features from the predicted bounding box for further feature updating. We use conditional inverse distance weighting to aggregate the features $\boldsymbol{q}_{j}$ regressed from these neighboring points to obtain a single updated feature $\boldsymbol{q}_{i}$ at updated proposal point $\boldsymbol{p}_{i}^{\prime}$:
\begin{equation}
    \boldsymbol{q}_{i}=\sum_{i} \mathbbm{1}_{i, j} \frac{w_{i}}{\sum w_{i}} \boldsymbol{q}_{j}, \text { where } w_{i}= -\exp{(\left\|\boldsymbol{p}^{\prime}_{i}-\boldsymbol{p}_{j}\right\|)},
\end{equation}
where $\mathbbm{1}_{i, j}$ is an indicator in the boundary mask that indicates whether proposal point $\boldsymbol{p}_{j}$ is in the bounding box predicted from proposal point $\boldsymbol{p}_{i}$. 

\subsection{Cascade Positive Assignment}
\label{CPA}

\begin{figure}[t]
   \centering
    \subcaptionbox{\label{sub-baseline} Directly Decrease $\mu$}{\includegraphics[width = .5\linewidth]{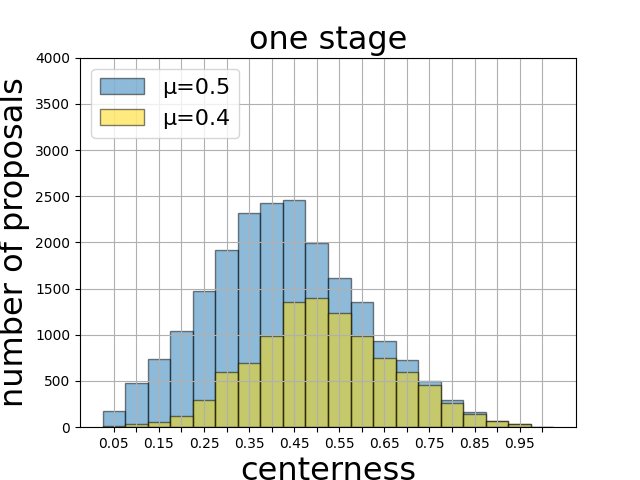}}\hfill
    \subcaptionbox{\label{sub-ours} Cascade Decrease $\mu$}{\includegraphics[width = .5\linewidth]{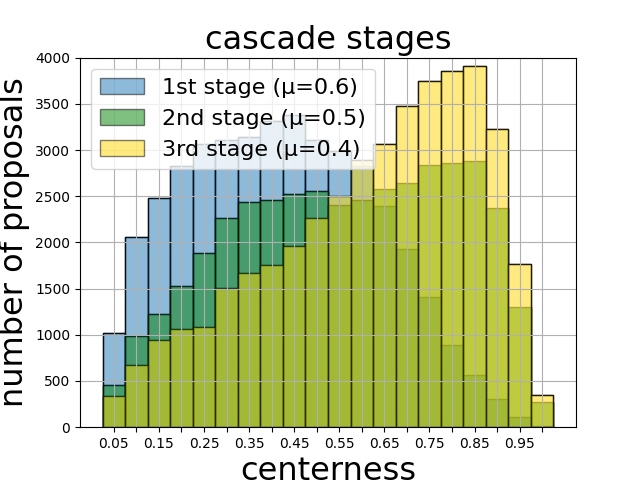}}
   \caption{(a) shows that the number of matched proposal points makes a sharp drop when directly the threshold for positives decreases from 0.5 (blue) to 0.4 (gold), which indicates that it is not feasible to reduce the threshold directly to remove noise proposal points due to the lack of positives for training. (b) shows in the first stage of our proposed CPA strategy, more points have a chance to be matched as positives than (a) due to $\mu>0.5$, and the centerness of these positives is further improved for excepted higher IoU in 2$^{nd}$ and 3$^{rd}$ stages with IA-voting modules. Thus we can perform decreasing $\mu$ to set up the stricter assignment with a sufficient number of high-centerness positives in the deeper stages.}
   \label{fig:stage-centerness}
\end{figure}

In order to ensure the number of positives with high centerness, we implement the CPA strategy for training, which consists of a sequence of $L$ positive assignment stages for each group of predicted objects from each detection head as shown in Figure \ref{fig:stage-positive}. The CPA trains the cascade decoder stages with decreasing center scale thresholds $\mu$, which are sequentially more selective against close false positives:
\begin{equation}
    \mu_{l} = \mu_{max} - \frac{l}{L} (\mu_{max} - \mu_{min}),
\end{equation}
where $l$ is the index of the decoder stage and starts with one. We set $\mu_{max}=0.4$ and $\mu_{min}=0.2$ by default. A larger $\mu$ relies less on the proposal centerness and meanwhile ensures objects that are easily overlooked are associated with the query. Thus the decoder relaxes the dependence on the proposal point centerness in the early stages as proposal points far from the ground truth centers also have a chance to be assigned as positives. Meanwhile, it ensures close enough points with higher centerness in the later stages with a more strict threshold as shown in Figure \ref{fig:stage-centerness}.

In addition, the proposal formulation still brings an instability problem where the targets may not be covered by the selected coarse proposal points due to unreliable predicted centerness. To mitigate this problem, we also leverage a denoising task as a training shortcut to avoid mismatched targets at the initial stage of training. Since we formulate each object feature from a 3D point, a noised point can be regarded as a “good” proposal that has a corresponding ground truth center nearby. The denoising training thus has a clear optimization goal - to predict the original bounding box from its nearby noised point. 

Specifically, we concat another point group as denoising points and reassign the first $T$ object features as denoising features, where $T$ is the number of ground truth boxes. Then we set $\boldsymbol{q}_{t}=\boldsymbol{x}^p_{k}$ and:
\begin{equation}
    k=\arg \min [\mathcal{D}_{\operatorname{match}}\left(\boldsymbol{p}_{n}, \boldsymbol{g}_{t}\right)]^{N}_{n=1},
\end{equation}
where $k$ is the index of denoising point and denoising feature, $\boldsymbol{x}^p\in\mathbb{R}^{N\times D}$ is the overall point features from the point encoder, $\mathcal{D}_{\operatorname{match}}$ is the $\ell_{1}$ distance metric, and $\boldsymbol{g}_{t}$ is the center point coordinate of the $t^{th}$ ground truth. This is equivalent to adding noise to the ground truth centers with a minimum distance from the proposal point, which is then gradually removed in the cascade stages of the decoder by the updating and refinement process. 

For training, each denoising point has a fixed assignment to the corresponding ground truth in the later stages, even though it has an updated proposal point at each stage, to explicitly enhance and guide the denoising process.

\begin{figure}[t]
   \centering
   \includegraphics[width=0.37\textwidth]{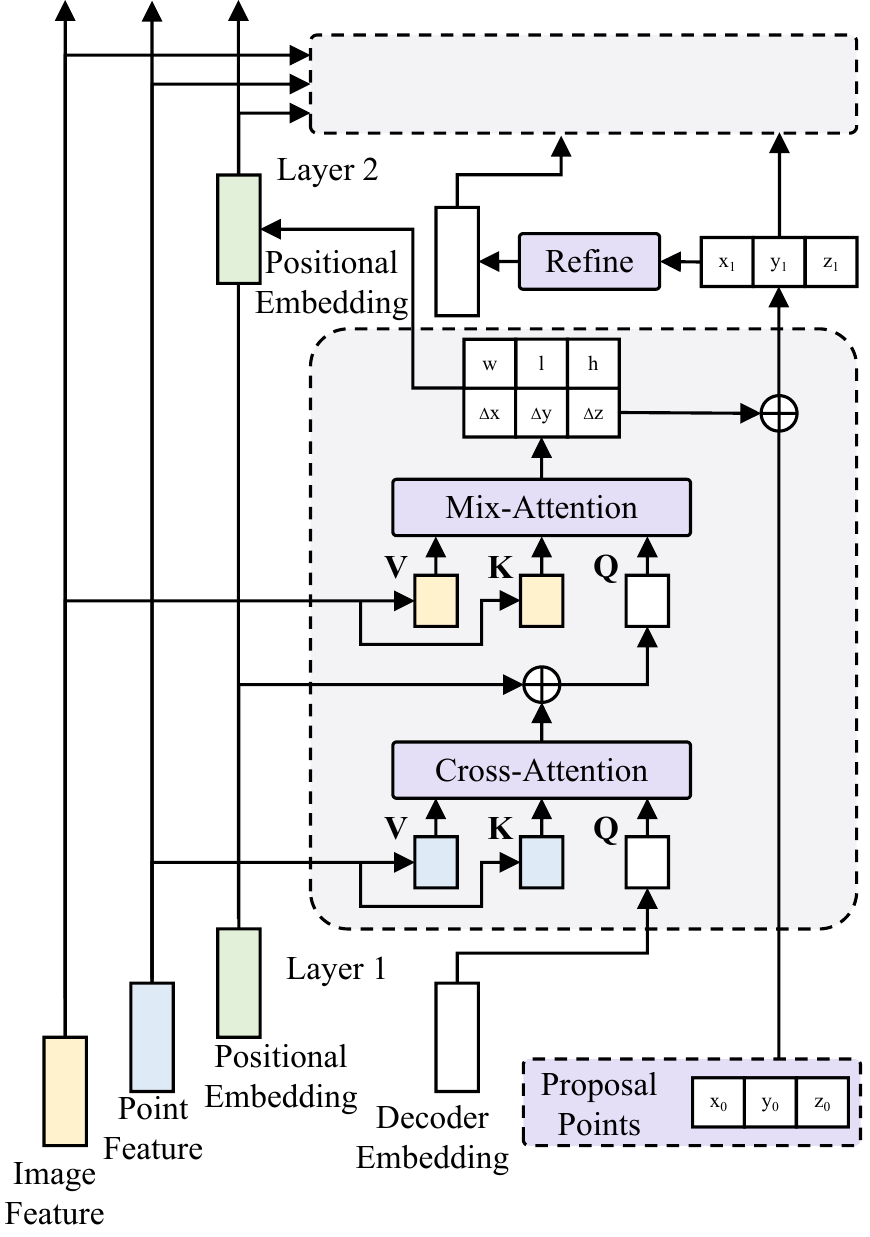}
   \caption{Details of our proposed transformer layer where the structure of each layer remains no change and only the attention part and refinement part are shown for clarity.}
   \label{fig:decoder-layer}
\end{figure}

\subsection{Fusing Image Features}
\label{img}

In order to utilize image features to assist 3D object detection, we re-design the transformer layers to flexibly support both pure point cloud input and optional point-image pair input. Specifically, we apply a well-trained Deformable-DETR encoder \cite{zhu2020deformable} on the RGB data as the image encoder in our method. Given an input image $\boldsymbol{I}\in\mathbb{R}^{H_0\times W_0\times 3}$, the image decoder is applied to extract multi-scale image embedding $\boldsymbol{x}^{i} \in \mathbb{R}^{M \times D}$, where $D=256$ is the feature dimension and $M$ is the sequence length.

With extracted point and image features, the transformer layers are applied afterward to supplement point features with semantic information extracted from the image as shown in Figure \ref{fig:decoder-layer}. The cross-attention module and the deformable attention module take the point cloud features $\boldsymbol{x}^{p}$ and image features $\boldsymbol{x}^{i}$ as keys and values, respectively, and optimize the query corresponding to the proposal point. Formally, one step of query optimization can be written as:
\begin{equation}
    \begin{array}{l}
    \boldsymbol{q}=\boldsymbol{q}+\operatorname{Cross-Attention}\left(\boldsymbol{q},\boldsymbol{x}^{p}\right), \\
    \boldsymbol{q}=\boldsymbol{q}+\operatorname{Deformable-Attention}\left(\boldsymbol{q},\boldsymbol{x}^{i}\right), \\
    \boldsymbol{q}=\boldsymbol{q}+\operatorname{FFN}\left(\boldsymbol{q}\right),
    \end{array}
\end{equation}
where reference points in deformable attention are mapped coordinate of 3D points in 2D images as \cite{yang2022boosting} does:
\begin{equation}
    \Psi(\hat{x}, \hat{y}, \hat{z})=\left(\frac{\psi_{1} \hat{x}+\psi_{2} \hat{y}+\psi_{3} \hat{z}}{\psi_{7} \hat{x}+\psi_{8} \hat{y}+\psi_{9} \hat{z}}, \frac{\psi_{4} \hat{x}+\psi_{5} \hat{y}+\psi_{6} \hat{z}}{\psi_{7} \hat{x}+\psi_{8} \hat{y}+\psi_{9} \hat{z}}\right),
\end{equation}
where $\psi$ are the parameters of the mapping function $\Psi$ related to the type of capture device and sensors.

\begin{table*}
  \centering
  \begin{tabular}{@{}lclclclclclclclc@{}lclclclclclclclc@{}lclclclclclclclc@{}lclclclclclclclc@{}lclclclclclclclc@{}lclclclclclclclc@{}lclclclclclclclc@{}lclclclclclclclc@{}lclclclclclclclc@{}lclclclclclclclc@{}lclclclclclclclc@{}}
    \toprule
    Methods & Input  & bed & table & sofa & chair & toilet & desk & dresser & nstand & bkshf & bathtub & average \\
    \midrule
    ImVoteNet \cite{qi2020imvotenet} & P+I & 87.6 & 51.1 &  70.7 & 76.7 & 90.5 & 28.7 & 41.4 & 69.9 & 41.3 & 75.9 & 64.4 \\
    DeMF \cite{yang2022boosting} & P+I & 89.5 & 53.5 & 75.0 & 80.7 & \textbf{94.3} & 36.3 & \textbf{50.3} & 72.4 & 43.0 & 79.5 & 67.4 \\
    FCAF3D+CascadeV & P+I & \textbf{90.4} & \textbf{58.5} & \textbf{77.8} & \textbf{83.4} & 93.1 & \textbf{40.6} & 50.1 & \textbf{73.5} & \textbf{48.9} & \textbf{85.8} & \textbf{70.2} \\
    \bottomrule
  \end{tabular}
  \caption{mAP@0.25 on SUN RGB-D where P indicates point cloud input and I indicates the corresponding RGB image input. The reported metric is average precision with an IoU threshold of 0.25 for per-class evaluation.}
  \label{tab:results-sunrgbd-perclass}
\end{table*}

\section{Experiments}

In this section, we evaluate the effectiveness of CascadeV-Det
on various datasets and with multiple evaluation metrics. Then we carry out an ablation study and extensive experiments with visualization on our proposed components to further prove their effectiveness. 

\subsection{Experimental Settings}

\noindent \textbf{Datasets.} We evaluate our method on the popular datasets of SUN RGB-D \cite{song2015sun} and ScanNet \cite{dai2017scannet}. SUN RGB-D includes 5k training scenes with bounding boxes in 10 classes, which are used to validate CascadeV-Det with both point-only and point-image input. To feed the data to our network, we first convert the depth images to RGB images and point clouds using the given camera parameters. The ScanNet dataset contains 1513 reconstructed 3D indoor scans with per-point instance labels of 18 object categories. The model is trained with only point cloud input on ScanNet. 


\noindent \textbf{Metrics.} The evaluation metric is the mean Average Precision (mAP) with the intersection over union (IoU) 0.25 (mAP@0.25) and 0.5 (mAP@0.5) same as VoteNet. We also report the evaluations on the 10 most frequent categories for SUN RGB-D. For relatively more stable results, We run training for 5 times and test each trained model 5 times independently with different random seeds. We report both the best and average performances across all 5 × 5 trials by default.

\noindent \textbf{Implement Details.} We implement our method using a single set aggregation operation to sample $B=512$ proposal points from overall $N=100,000$ points. We obtain 256-dimensional point and image features. We use iterative object box prediction and the loss functions in the proposal and classification stage that consist of objectness, bounding box estimation, and semantic classification losses following \cite{rukhovich2021fcaf3d}. The coarse network and CascadeV-Det decoder are trained and optimized separately using the AdamW optimizer for 36 epochs with the step learning rate decayed by 0.1 at 24$^{th}$ and 32$^{th}$ epoch. A weight decay of 0.01 is also used. We train the model on two GeForce RTX 3090 GPUs with a batch size of 16. We use the RandomFlip only for data augmentation.

\begin{table}
  \centering
  \begin{tabular}{@{}lclclc@{}lclclc@{}lclclc@{}}
    \toprule
    Method & Input & mAP@0.25 & mAP@0.5 \\
    \midrule
    VoteNet \cite{qi2019deep} & P & 60.0 & 41.3 \\
    Group-free \cite{liu2021group} & P & 63.0 (62.6) & 45.2 (44.4) \\
    TokenFusion \cite{wang2022multimodal} & P & 64.9 (64.4) & 48.3 (47.7) \\
    FCAF3D \cite{rukhovich2021fcaf3d} & P & 64.2 (63.8) & 48.9 (48.2) \\
    FCAF3D+CascadeV & P & \textbf{65.6} (65.0) & \textbf{49.2} (48.7) \\ \hline
    ImVoteNet \cite{qi2020imvotenet} & P+I & 64.4 & 43.3 \\ 
    DeMF \cite{yang2022boosting} & P+I & 67.4 (67.1) & 51.2 (50.5) \\
    FCAF3D+CascadeV & P+I & \textbf{70.2} (69.6) & \textbf{51.4} (50.8) \\
    \bottomrule
  \end{tabular}
  \caption{mAP on SUN RGB-D dataset. The reported metric value is the best one across 5 × 5 trials and the average value is also given in brackets. The results of the comparison methods are from the respective original papers.}
  \label{tab:results-sunrgbd}
\end{table}

\begin{table}
  \centering
  \begin{tabular}{@{}lclclc@{}lclclc@{}lclclc@{}}
    \toprule
    Method & Input & mAP@0.25 & mAP@0.5 \\
    \midrule
    VoteNet \cite{qi2019deep} & P & 58.6 & 33.5 \\
    Group-free \cite{liu2021group} & P & 69.1 (68.6) & 52.8 (51.8) \\
    TokenFusion \cite{wang2022multimodal} & P & 70.8 (69.8) & 54.2 (53.6) \\
    FCAF3D \cite{rukhovich2021fcaf3d} & P & 71.5 (70.7) & 57.3 (56.0) \\
    FCAF3D+CascadeV & P & \textbf{72.4} (72.0) & \textbf{57.7} (56.8) \\
    \bottomrule
  \end{tabular}
  \caption{Results on ScanNet dataset with point-only input where the point encoder in CascadeV-Det is FCAF3D. We only compare the results with the point cloud input on ScanNet as it involves multiple 2D views and requires extra handling to merge multi-view features.}
  \label{tab:results-scannet}
\end{table}


\subsection{Comparing with State-of-the-art Methods}

 First, we compare the performance of CascadeV-Det and methods with both image and point input. Table \ref{tab:results-sunrgbd-perclass} shows the per-class results on SUN RGB-D for methods with both point and image input. Our method gets better results on nearly all categories and has the biggest improvements on object categories whose points are more densely distributed on the surface. For example, CascadeV-Det achieved a better +4.3\% and +5.9\% improvement than the state-of-the-art method on the most difficult-to-detect desk and bookshelf. Second, we further report the performance of CascadeV-Det with and without image reference. As shown in Table \ref{tab:results-sunrgbd}, when only the point cloud is used as input, our modules achieve a boost of +1.4\% mAP@0.25 and 0.3\% mAP@0.5 over the original FCAF3D, which outperforms all previous fusion methods by a large margin. Particularly, with the image features, it surpasses the current best method by +2.8\% on mAP@0.25 and +0.2\% on mAP@0.5, demonstrating the superiority of our method. Table \ref{tab:results-scannet} also reports the results on ScanNet and S3DIS with point input only.


\subsection{Ablation Study}

In this section, we run ablation experiments on SUN RGB-D with a varying number of stages, the number of $B$ queries selected by the decoder, and with and without Cascade Voting decoder (CDeN) or Instance Aware Voting (IA-Voting). We also discuss the design choices of CascadeV-Det and investigate how they affect the evaluation metrics when applied independently. 

\noindent \textbf{Effectiveness of IA-Voting.} To validate the effectiveness of our proposed IA-Voting module, we build a model with the point encoder, one-stage predictions, and standard positive assignment with $\mu=0.5$ as the base model. The result of the base model with the IA-Voting module with an extra stage is available in Table \ref{tab:results-sunrgbd-ablations}. We observe that the mAP@0.25 benefits the most from the IA-Voting module, bringing a remarkable increment of 1.2\%. This indicates that IA-Voting can effectively reduce the problem of missing detection with the too-small IoU between the predicted bboxes and ground truth caused by coarse proposal points.

\noindent \textbf{Effectiveness of CPA.} Table \ref{tab:results-sunrgbd-ablations} also shows the results on the model trained with the CPA strategy, where the model trained with the CPA strategy and IA-Voting achieves remarkable gains of +1.6\% and +0.8\% in terms of mAP@0.25 and mAP@0.5. The results show that the model trained with the standard positive assignment by ensuring the high centerness of the positives.

\begin{figure}[t]
    \centering
    \subcaptionbox{\label{6}}{\includegraphics[width = .48\linewidth]{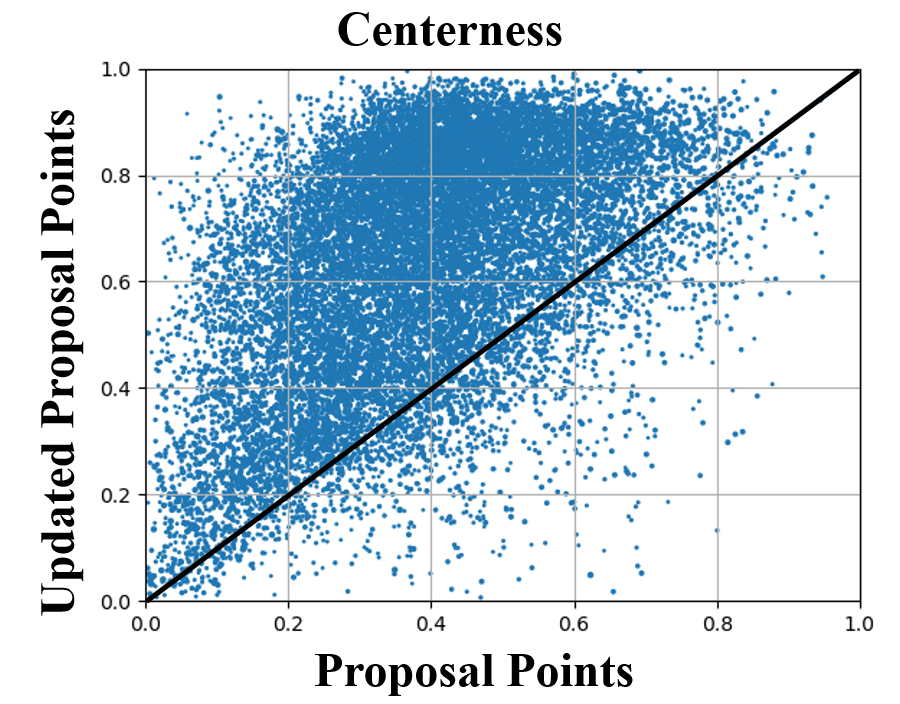}}
    \subcaptionbox{\label{5}}{\includegraphics[width = .48\linewidth]{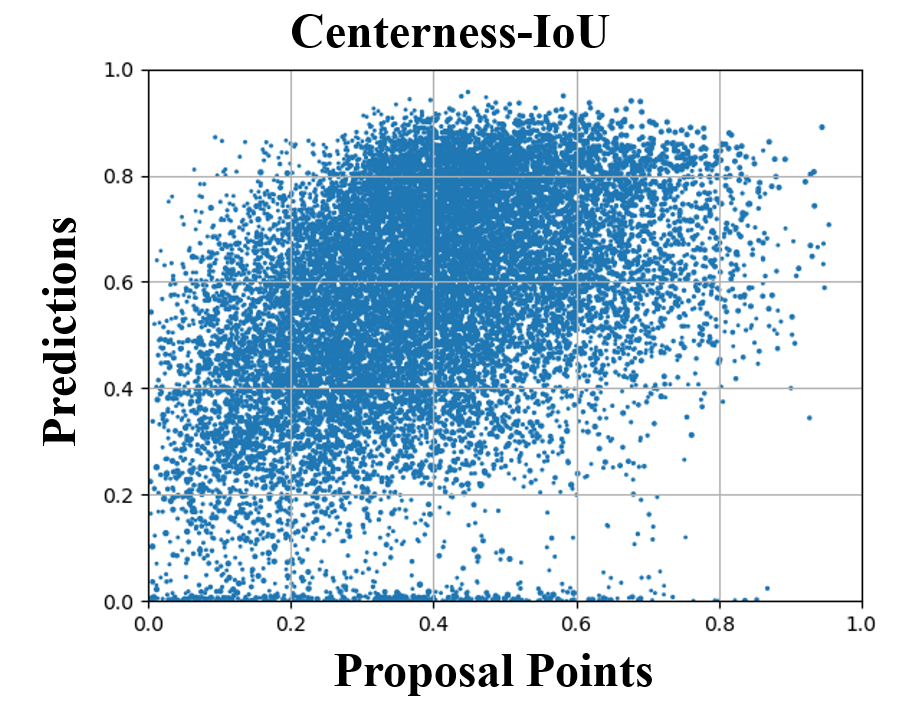}}\hfill
    \caption{(a) shows that the updated proposal points tend to have larger centerness than the original proposal points since most drawn points are above the line with slope one. This implies that they work as more accurate proposal points for the next stage. (b) indicates that predicted bounding boxes with larger IoU are expected from proposal points with higher centerness.}
    \label{fig:centerness}
\end{figure}

\begin{table}
  \centering
  \begin{tabular}{@{}lclclc@{}lclclc@{}}
    \toprule
    Method & mAP@0.25 & mAP@0.5 \\
    \midrule
    FCAF3D & 64.0 & 48.4 \\
    \quad+\  IA-Voting & 65.2 & 48.0 \\ 
    \quad+\  IA-Voting +\  CPA\  & \textbf{65.6} & \textbf{49.2} \\ 
    \bottomrule
  \end{tabular}
  \caption{We incrementally add the IA-Voting module and the CPA training strategy to the FCAF3D-based point encoder to verify the effectiveness of both new components.}
  \label{tab:results-sunrgbd-ablations}
\end{table}

\begin{table}
  \centering
  \begin{tabular}{@{}lclclc@{}lclclc@{}lclclc@{}}
    \toprule
    stages & test stage & mAP@0.25 & mAP@0.5 \\
    \midrule
    1 & 1 & 69.8 & 50.6 \\
    2 & $\underline{1 \sim 2}$ & 70.2 & 51.4 \\
    3 & $\underline{1 \sim 3}$ & \textbf{70.4} & \textbf{51.6} \\
    \bottomrule
  \end{tabular}
  \caption{The impact of the number of decoder layers. $\underline{i \sim j}$ indicates that the model ensembles the predicted bounding boxes from stage $i$ to stage $j$ with joint NMS to produce final detection results.} 
  \label{tab:results-sunrgbd-decoders}
\end{table}

\begin{figure*}[t]
   \centering
   \includegraphics[width=0.95\textwidth]{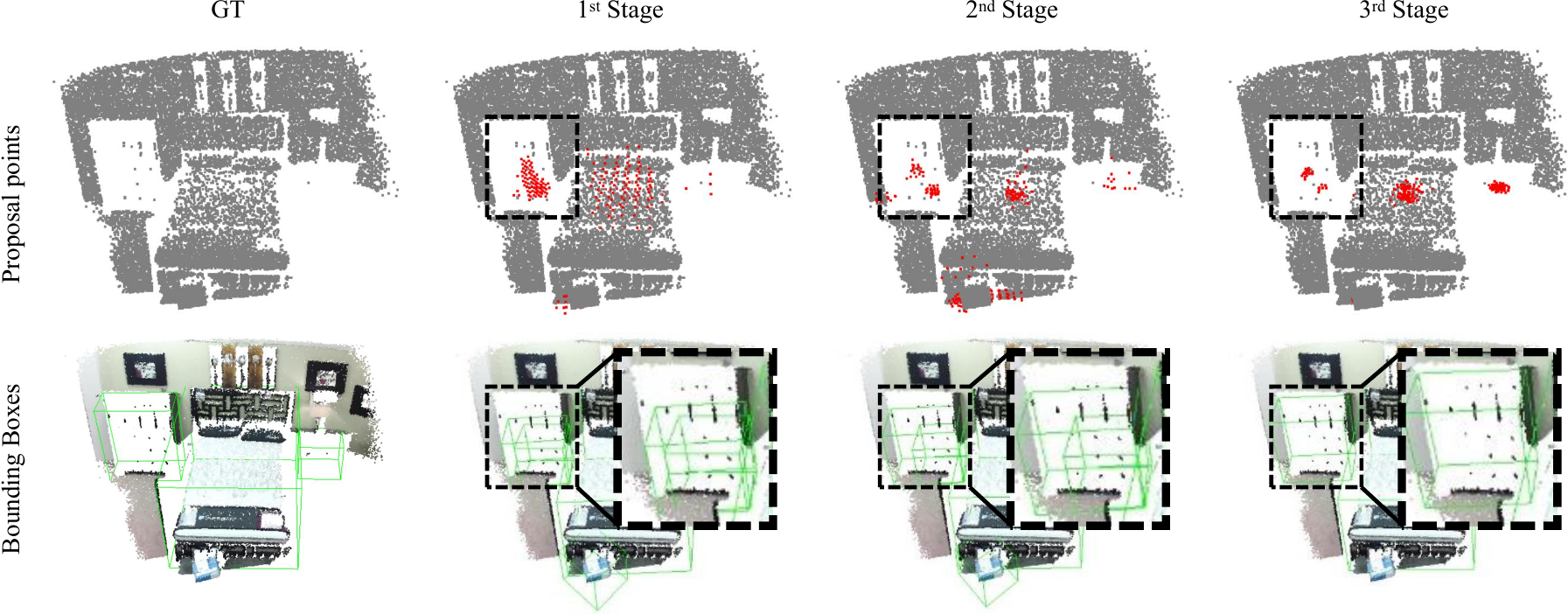}
   \caption{Qualitative results of proposal points and predicted bounding boxes from multiple stages on SUN RGB-D dataset. The proposal points are gathered at the corresponding ground truth centers and aggregate more accurate object features by IA-Voting modules. And the false positives are gradually eliminated because the CPA strategy applies a smaller positive assignment threshold in the later stage.}
   \label{fig:visualization-predictions}
\end{figure*}

\noindent \textbf{Number of Cascade Stages.} Table \ref{tab:results-sunrgbd-decoders} summarizes stage performance. The models with cascade stages all outperform the baseline, due to the benefits of updating process and cascade training. The following stages improve performance substantially and a detector with 3 stages provides the best performance with the ensembles of all detection stages.

\subsection{Qualitative Results and Discussion}

In this section, we verify our two basic assumptions by statistical results from the input proposed points and their corresponding predictions: 1) The updated proposal points are closer to the ground truth centers than the original proposal points and in other words have higher centerness. 2) The predicted bounding boxes gain larger IoU from proposal points with higher centerness. As shown in Figure \ref{fig:centerness}, we calculate the centerness of the proposal points in all stages from CascadeV-Det and show the centerness of the corresponding centerness of the updated proposal points and the IoU of predicted bounding boxes. The results show the correctness of our assumptions and the effectiveness of our method, providing strong support for the proposed cascade updating manner.

In Figures \ref{fig:visualization-predictions}, we visualize the updated proposal points of each stage from a model trained with 3 cascade stages. We observe that some proposal points (colored) are clustered near the ground truth centers and other samples locate around the boundary of the objects. Then the discrete proposal points are cluttered in the later stages by Cascade Voting, which compensates for the distance from surface points to the ground truth centers and shows how IA-Voting can help 3D detection by aggregating instance-level object features from updated proposal points. The predicted bounding boxes also show that the CPA strategy trains the decoder well to avoid false positives.

\section{Conclusion}

In this paper, we introduce CascadeV-Det for 3D object detection from point clouds to provide high quality 3D object detection performing point-based prediction with two novel components: the IA-Voting module and the CPA strategy. Statistic and visualization analysis shows the effectiveness of both the new components. Our proposed CascadeV-Det significantly outperforms the previous state-of-the-art on the challenging SUN RGB-D and achieves competitive results on the ScanNet benchmark in terms of mAP. In the future, we will further investigate the more challenging wild scenes and expand our method to other 3D scene understanding tasks.



{\small
\bibliographystyle{ieee_fullname}
\bibliography{egbib}
}

\end{document}